\documentclass[journal]{IEEEtran}

\ifCLASSINFOpdf
   \usepackage[pdftex]{graphicx}
   \graphicspath{{./jpeg/}}
   \DeclareGraphicsExtensions{.pdf,.jpeg,.png}
\else

\fi

\usepackage{amsmath,bm}
\usepackage{booktabs,caption}
\usepackage[flushleft]{threeparttable}
\usepackage[shortlabels]{enumitem}
\usepackage{enumitem}
\usepackage{rotating}
\usepackage{url}
\usepackage{cite}

\usepackage{breakurl}
\usepackage[breaklinks]{hyperref}
\usepackage{tikz}
\def\checkmark{\tikz\fill[scale=0.4](0,.35) -- (.25,0) -- (1,.7) -- (.25,.15) -- cycle;} 
\usepackage{amssymb} 
\usepackage{tabularx, booktabs, multirow}
\newcolumntype{C}{>{\centering\arraybackslash}X} 
\begin{document}

\title{ Early Lane Change Prediction for Automated Driving Systems Using Multi-Task Attention-based Convolutional Neural Networks}

\author{Sajjad~Mozaffari,
			Eduardo~Arnold,
			Merhdad~Dianati,
			
			and~Saber~Fallah
\thanks{This  work  was  supported  by  Jaguar  Land  Rover  and  the  U.K.-EPSRC  as part of the jointly funded Towards Autonomy: Smart and Connected Control(TASCC) Programme under Grant EP/N01300X/1}
\thanks{S. Mozaffari, E.Arnold, and M. Dianati are with the Warwick Manufacturing Group, University of Warwick, Coventry CV4 7AL, U.K. (e-mail: {sajjad.mozaffari, e.arnold, m.dianati}@warwick.ac.uk)}
\thanks{S. Fallah is with the Department of Mechanical Engineering Sciences, University of Surrey, Guildford, GU2 7XH, U.K. (e-mail: s.fallah@surrey.ac.uk}
}

\maketitle

\begin{abstract}
Lane change (LC) is one of the safety-critical manoeuvres in highway driving according to various road accident records. Thus, reliably predicting such manoeuvre in advance is critical for the safe and comfortable operation of automated driving systems. The majority of previous studies rely on detecting a manoeuvre that has been already started, rather than predicting the manoeuvre in advance. Furthermore, most of the previous works do not estimate the key timings of the manoeuvre (e.g., crossing time), which can actually yield more useful information for the decision making in the ego vehicle. To address these shortcomings, this paper proposes a novel multi-task model to simultaneously estimate the likelihood of LC manoeuvres and the time-to-lane-change (TTLC). In both tasks, an attention-based convolutional neural network (CNN) is used as a shared feature extractor from a bird's eye view representation of the driving environment. The spatial attention used in the CNN model improves the feature extraction process by focusing on the most relevant areas of the surrounding environment. In addition, two novel curriculum learning schemes are employed to train the proposed approach. The extensive evaluation and comparative analysis of the proposed method in existing benchmark datasets show that the proposed method outperforms state-of-the-art LC prediction models, particularly considering long-term prediction performance.

\end{abstract}

\begin{IEEEkeywords}
Vehicle Behaviour Prediction, Automated Driving, Multi-task Learning, Curriculum Learning, Attention Mechanism
\end{IEEEkeywords}

\IEEEpeerreviewmaketitle

\section{Introduction} \label{intro}


\IEEEPARstart{A}{nticipating} the future behaviour of surrounding vehicles is a critical function of both Advanced Driving Assistance Systems (ADAS) and fully automated driving systems. Failing to do so may lead to either hazardous driving decisions or forcing unduly conservative driving, compromising the safety or efficiency of the driving system, respectively. 

In highway driving, lane change (LC) manoeuvres are considered to be safety-critical as they can lead to hazardous interference with the traffic in other lanes. A major crash dataset collected between 2010 and 2017 in the United Arab Emirates indicates that accidents associated with unsafe LC manoeuvres are among the main causes of severe injuries in highway driving~\cite{Shawky2020}. Although monitoring the indicator signals can be a good detector of  LC manoeuvres, studies in USA~\cite{LC_signal_use_USA} and China~\cite{LC_signal_use_China} show that less than $50\%$ of drivers are using signal indicators for doing an LC manoeuvre. The LC manoeuvres that occur without proper signalling are often the culprit in accidents. Such hazards can be detected by analysing the motion of the vehicles, which is the main focus of this paper. A reliable LC prediction model in such cases can provide an early warning of emerging LC manoeuvres of other vehicles. Such information can then be used by the ego-driver or the automated driving system to pro-actively make driving decisions and alleviate the risk of accidents associated with LC manoeuvres. An early manoeuvre prediction has a particularly higher value in highway driving scenarios, where the high-speed of vehicles require a more agile driving style and decision making.

\begin{figure}[!t]
\centering
\includegraphics[width=3.4in]{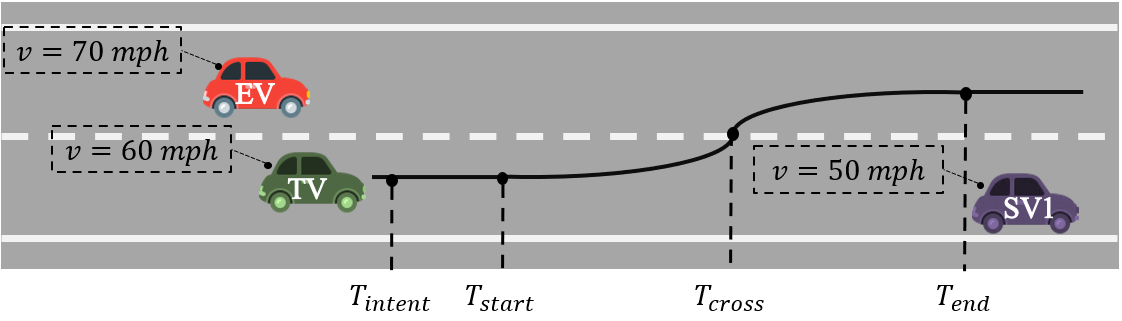}
\caption{An example of an LC scenario in a left-hand driving road system. Target Vehicle (TV): Green Car, Ego Vehicle(EV): Red Car, and Surrounding Vehicle 1 (SV1): Purple Car
}
\label{example_fig}
\end{figure}

An example of an LC scenario in a highway driving environment is illustrated in Figure~\ref{example_fig}. In this scenario, the Target Vehicle (TV) decides to perform a lane change manoeuvre at time $T_{intent}$ due to the slow-moving preceding vehicle (i.e., SV1: Surrounding Vehicle 1).  The LC manoeuvre starts at $T_{start}$, as the vehicle drifts towards the left lane marking. The TV crosses the lane marking at $T_{cross}$ and finishes its LC manoeuvre by stabilizing its lateral position at the centre of the left lane at $T_{end}$. The Ego vehicle (EV), which is already driving on the left lane, is supposed to decelerate as soon as it realises the imminent LC manoeuvre by the TV. Therefore, the EV needs to have an early and reliable prediction of the LC manoeuvre and its key timings (e.g., $T_{cross}$) to perform a smooth and safe deceleration.

There have been several studies on LC prediction in the literature. A major group of existing studies has shown to be able to predict a lane crossing that occurs up to 2.5 seconds in the future~\cite{Manttari2018, Llorca2020}, while an LC manoeuvre, from $T_{start}$ to $T_{end}$, usually takes between 3 to 5 seconds~\cite{Toledo2007}. This means that most existing studies can only predict an LC manoeuvre after $T_{start}$, i.e., after the manoeuvre has already started.  These studies mostly rely on detecting the lateral drift of the TV towards the lane marking. However, there are fewer clues in the past motion of the TV for predicting its future manoeuvre in longer horizons. Therefore, long-term prediction approaches need to understand the traffic context around the TV, rather than solely analysing the recent motion of the TV. In recent years, some studies attempted to extend the prediction horizon by using deep learning-based models, mainly Long Short-Term Memories (LSTMs)~\cite{Ding2019, Wirthmuller2021}. However, the use of LSTMs for long-term LC prediction is impeded by their shortcoming in extracting spatial interdependency which is required to model the interaction among nearby traffic agents.

In most existing studies~\cite{Izquierdo2019
,Ding2019, Mahajan2020, Llorca2020, Hu2018, Lee2017} the problem of LC prediction is defined as predicting the likelihood of LC manoeuvres over the next few seconds. In such a formulation, an LC manoeuvre of the TV occurring early in the future is treated equally as an LC manoeuvre happening farther in time. As a result, such an approach cannot inform the EV about the key timings of the TV's manoeuvre, which is crucial for safe and comfortable trajectory planning in automated vehicles.

In this paper, we address the aforementioned shortcomings by proposing a multi-task attention-based prediction model. We apply a novel Convolutional Neural Network (CNN) with spatial attention to a bird's eye view representation of the traffic context around the TV to extract relevant features. An attention module is employed to selectively focus on the most informative areas of the TV's surroundings to improve the feature extraction process. The features extracted by the attention-based CNN are then used to simultaneously predict the likelihood of LC manoeuvre and the time-to-lane-change in a Multi-Task Learning (MTL) approach. Also, two Curriculum Learning (CL) criteria are introduced during the training phase to increase the generalisation of the proposed model. The proposed joint LC and TTLC prediction model is trained and evaluated using a public large scale trajectory dataset collected from German highways~\cite{highDdataset}. The contributions of this work are summarised below:
\begin{itemize}
\item A novel multi-task formulation of lane change prediction to predict both the type and timing of lane change manoeuvre.
\item A novel spatial attention module on top of CNN-based feature extractor to selectively focus on informative areas around target vehicle.
\item Comprehensive evaluation of the proposed model and some state-of-the-art prediction models in terms of prediction performance and horizon in both regression and classification formulation.
\end{itemize}

The rest of this paper is organised as follows. Section~\ref{rel_works} reviews recent LC prediction approaches based on their input representation, prediction model, and output type. Section~\ref{prob_form} discusses the system model and problem formulation. Section~\ref{prop_method} introduces the proposed method and its key components. Section~\ref{exps} presents the experiments' setup for evaluating the performance of the proposed method, key results from those evaluations and the discussions of the insights that can be learned from our performance evaluation. Finally, some key concluding remarks are given in section~\ref{conclusion}.

\section{Related Works} \label{rel_works}
Overviews of vehicle behaviour prediction approaches are presented in~\cite{Lefevre2014, Mozaffari2020}. Vehicle behaviour prediction studies can be divided based on output type into trajectory prediction~\cite{Greer2021, koschi2021, Messaoud2021} and manoeuvre prediction~\cite{Zyner2020, Zhang2018, Bahram2016}. A trajectory prediction model attempts to predict the continuous states of a vehicle (e.g., x-y location) for each timestep during a prediction window, while a manoeuvre prediction model anticipates the type of manoeuvre a vehicle is intended to perform. As a sub-category of manoeuvre prediction, lane change prediction has been widely studied in recent years. Studies on lane change prediction can be further divided into ego vehicle lane change prediction (also known as driver lane change inference) and surrounding vehicle lane change prediction, which are separately reviewed in~\cite{Xing2019} and~\cite{Song2021}, respectively. The focus of this paper is on lane change manoeuvre prediction of surrounding vehicles in highway driving scenarios. In this section, we review lane change manoeuvre prediction studies based on their input representation and prediction model and we highlight how our paper is differentiated from them.  







\subsection{Input representation}
Different input data can provide clues for an upcoming lane change manoeuvre. The inputs can be categorised as follow:
\begin{enumerate}
\item \textbf{TV's States:} The state of the TV in recent time-steps contains the main clue for a lane change manoeuvre happening in near future. Therefore, most lane change detection or short-term prediction rely only on this type of feature. TV's lateral position in the lane, lateral and longitudinal velocity and acceleration are examples of TV states used as an input in previous studies~\cite{Mahajan2020, Yoon2016, Liu2019}.

\item \textbf{Environment States:} In early lane change prediction, there are not enough information in recent TV's states since this is the time that the driver is planning to perform a lane change. The intention of a lane change can form based on an internal driving goal (e.g., reaching an exit on a highway) or static environmental conditions (e.g., ending of a lane in merging scenario) or dynamic environment conditions (e.g., a slow-moving preceding vehicle). Although it is usually not possible to observe the internal driving goal of a vehicle's driver, automated vehicles can observe the driving environment through perception sensors, V2V/V2I communication and HD maps. Therefore TV's states can be augmented with a representation of static and dynamic environment states for longer-term predictions. In~\cite{Rehder2019, Scheel2019, Ding2019, Hu2018, Rehder2016} a list of interaction-aware features like relative distance to surrounding vehicles, relative velocity, and so on has been used to model the interaction between the TV and surrounding vehicles. Some of these studies also contain features describing the driving environment such as distance to the nearest on- or off-ramp \cite{Scheel2019} and the existence of the lanes \cite{Rehder2019, Wirthmuller2021}. In~\cite{Shou2020}, the features describing the lateral motion of vehicles have been removed from input representation to predict very long (5 to 10 seconds) left lane change manoeuvres. They argue that lateral features are not informative in very long prediction horizons. Some of the existing studies utilize Deep Neural Networks (DNNs) to learn relevant features from raw sensor data~\cite{Llorca2020, Izquierdo2019}. Although such a strategy leads to no information loss, large computational resources are required to learn relevant features from high-dimensional raw sensor data, which is challenging due to the limited computational resources of an automated vehicle. Therefore, some studies try to learn features from a simplified bird eye view (BEV) representation of the driving environment instead of raw sensor data~\cite{Cui2019, Deo2018, Lee2017, Manttari2018}. This representation depicts vehicles as their bounding boxes and the road markings and it has a significantly lower dimensionality compared to raw sensor data. Although a BEV representation has a larger dimension compared to using a list of interaction-aware features, it has been shown that representing TV and surrounding vehicles states as well as the static driving environment in a BEV image representation can facilitate joint feature learning for vehicle behaviour prediction using deep neural networks~\cite{Cui2019, Deo2018, Lee2017, Manttari2018}. In this paper, we adopt the same approach and design a simplified BEV input representation for the problem of lane change prediction. Section \ref{in_rep} explains the details of the simplified BEV representation used in our study.

\item \textbf{Driver's States:} Driver's states such as head position and gaze movement are usually used when the behaviour of ego vehicle is being predicted~\cite{Yan2019, Khairdoost2020}. Such features are usually not easily observable in the case of surrounding vehicle lane change prediction. Therefore, the driver's states are not considered in our input representation. 
\end{enumerate}

\subsection{Prediction Model}

A group of existing studies utilise graphical models to predict LC manoeuvre. Graphical models are probabilistic models for which a graph, defined using domain knowledge, expresses the conditional dependencies among input, output, and hidden random variables.
Hidden Markov Models (HMMs), as a subclass of graphical models, have been widely used in behaviour recognition and prediction~\cite{Deng2021}. In~\cite{Zhang2018} two continuous HMMs are used to model lane keeping and lane changing behaviour separately. The model in~\cite{Zhang2018} has been trained and tested on the NGSIM I-80 dataset with a recall rate around $85\%$ in 0.5 seconds and $65\%$ in 4 seconds prediction horizons.
 In \cite{Rehder2016}, a Bayesian network is used to compute the driver's contentedness (i.e., likelihood of occupancy) for all lanes the driver could drive. The driver's contentedness for each lane is then employed to predict the LC intentions. Their approach relies on lane change intention labels extracted from human driving in a simulator environment.
In~\cite{Bahram2016} a multi-agent simulation with model-based loss function for interaction modelling and a Bayesian network classifier has been used as a combined model-based and learning-based approach. They have reported the lateral motion prediction results for prediction horizons up to 2.5 seconds.
 In \cite{Rehder2019}, a hybrid Dynamic Bayesian Network (DBN) is adopted to predict LC intention using a list of interaction-aware features. The latent variables of the prediction model are pre-trained using simulation data, followed by training on German highway driving data. Since exact inference is intractable in DBNs, time-series information is not considered in the inference process.  The Reciever Operating Characteristic (ROC) of the proposed approach has been reported for lane change occurring 3 seconds in the future.
 A three-layer DBN has been used in \cite{Liu2019} to estimate the driver LC intention and driver characteristic. Then, a Gaussian process is exploited to predict trajectory based on predicted LC manoeuvre. The authors of~\cite{Liu2019} reported achieving $95\%$ accuracy in LC prediction; however, the prediction horizon for this performance has not been reported. 
 
  The advantage of graphical models lies in their ability to interpret the model's prediction by examining the values of the graph nodes. However, the drawback of such an approach stems from their limited expression capabilities since no reasoning can be done beyond the fixed relation defined between the graph nodes. In addition, the exact inference in graphical models is often intractable. Therefore, most prediction approaches using graphical models are considered for short to medium prediction horizons up to 3 seconds.

Most recent works exploit DNNs for LC prediction. For example, fully connected feed-forward neural networks have been used in~\cite{Yoon2016, Hu2018, Kruger2019, Wirthmuller2020J}. Several existing studies apply variants of recurrent neural networks to predict LC manoeuvre \cite{Mahajan2020, Zou2019, Ding2019,  Scheel2019, Wirthmuller2021}, inspired by their success in data sequence analysis. In~\cite{Mahajan2020} and~\cite{Zou2019}, a single LSTM is applied to TV-only and interaction-based features for LC prediction, respectively. In \cite{Ding2019}, a model containing several Gated Recurrent Unit (GRU) is proposed to model pairwise interaction between the TV and each of the SVs. Convolutional Neural Networks (CNNs) have been used in some of the existing studies mainly to model spatio-temporal dependencies in image-like input data such as simplified BEV representation~\cite{Manttari2018, Lee2017, Deo2018} or raw sensor data~\cite{Llorca2020, Izquierdo2019}. In this paper, a convolutional neural network is also constructed to learn lane change related features from stacked BEV input representation. In~\cite{Scheel2019}, the input features are categorized into TV's motion, Right lane SVs, Left Lane SVs, Same lane SVs and Street features based on which part of the environment they represent. Then, an LSTM is used per feature group to create an internal representation of features. Finally, an attention mechanism is used to specify the importance of each group of features for LC prediction in each data sample. Attention mechanism has been also used in~\cite{Messaoud2021} to pay selective attention to a subset of surrounding vehicles in similar problem of trajectory prediction. Unlike~\cite{Scheel2019, Messaoud2021} where the attention mechanism is applied on each 1D vector of surrounding vehicles' states, we propose a spatial attention mechanism on top of 2D feature maps of convolutional neural networks. The proposed spatial attention mechanism selectively focusses on different quarters of the driving environment around the TV. Unlike previous studies, the traffic situation, including both static and dynamic contexts of each quarter simultaneously contribute to attention weights, which allows attending to free spaces as well as vehicles in each quarter. 

Although DNNs have shown promising performance in LC prediction, they suffer from a lack of interpretability, which negatively affects the social acceptance, debugging, and validation of such approaches. The proposed spatial attention mechanism in this paper can increase the interpretability of our deep learning-based approach by providing an interpretable intermediate representation in the form of attention weights of each quarter area of the surrounding environment.

There are a few studies that attempt to predict the timing of lane change manoeuvres~\cite{Wissing2017, Wirthmuller2021}. In~\cite{Wissing2017}, two quantile regression techniques, namely Linear Quantile Regression and Quantile Regression Forests have been compared in time to lane change prediction. Their approach relies on TV and environment states and provides probabilistic outputs. The results show that the Root Mean Square Error(RMSE) of prediction falls below 1 second only in prediction horizons below 1.5 seconds. In a recent study~\cite{Wirthmuller2021}, LSTM models have been thoroughly investigated for predicting the time until a vehicle changes the lane. The selected features are a list of interaction-aware features used in their previous study~\cite{Wirthmuller2020C} and the problem is defined as predicting the time to left lane change and the time to right lane change, separately.





\begin{figure*}[!t]
\centering
\includegraphics[width=6.8in]{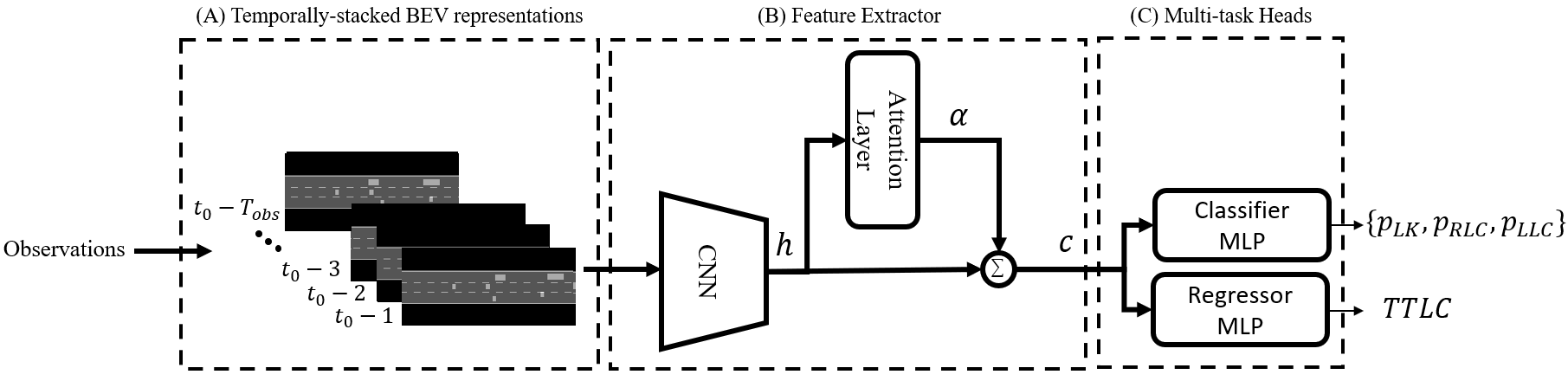}
\caption{An overview of key processing steps of the proposed method}
\label{whole_model}
\end{figure*}

\section{System Model and  Problem Definition} \label{prob_form}

We assume a semi- or fully automated EV aims to predict the lane change manoeuvre of a nearby vehicle, called the TV. Both EV and TV are driving on a straight highway with an arbitrary number of lanes. There is an arbitrary number of surrounding vehicles (SVs) driving in the vicinity of the TV. Similar to most existing studies on LC prediction~\cite{Ding2019,Mahajan2020, Yoon2016, Hu2018, Manttari2018, Deo2018, Scheel2019, Liu2019}, we assume a BEV camera installed on an infrastructure building or a drone is observing the driving environment, detecting the vehicles, and tracking them while driving on a road section. The vehicles' tracking history, as well as the position of road markings, are assumed to be shared with the EV for the problem of LC prediction. Note that the vehicles' tracking data can also be obtained from egocentric perception sensors (instead of infrastructure sensors) with less coverage and accuracy due to their limited range and occlusion.We refer readers to~\cite{Mozaffari2021} for a comparative study of different perception approaches for lane change prediction in highways.

We divide the problem of LC prediction into a classification and a regression sub-problem. The classification problem aims to estimate the probability of LC manoeuvres occurring during a prediction window, $T_{pw}$. The LC manoeuvres are categorized into  1) Right Lane Change (RLC), 2) Left Lane Change (LLC), and 3) Lane Keeping (LK). The value of $T_{pw}$ determines the maximum prediction horizon of the prediction model. In this study, we set $T_{pw} = 5.2\mbox{sec}$ which enables evaluating the prediction model for long prediction horizons while assures a sufficient number of data samples to be extracted from the selected dataset (refer to section~\ref{dataset}). The regression problem aims to estimate the Time To Lane Change (TTLC). The TTLC is defined as the shortest time until the centre of the TV crosses either left or right lane marking. Predicting TTLC is crucial since it is the first time the TV interferes with the traffic on adjacent lanes. The input data to both subproblems are the TV's and its surrounding vehicles' (SVs) states and the position of lane markings. during a temporal observation window of 2 seconds, $T_{obs} = 2  \mbox{ sec}$.

\section{Proposed Method}
\label{prop_method}

This section presents the proposed multi-task attention-based LC prediction model and the curriculum learning schemes used in the training of the model. The processing steps in the proposed prediction model are summarised as follows:

\begin{enumerate}[(A)]
\item The observation of the TV and its SVs during $T_{obs}$ are used to render a simplified BEV representation of the driving environment for each time-step of $T_{obs}$.
\item An attention-based CNN is applied to the temporally stacked BEV representations to extract informative features for LC prediction.
\item The features extracted from the previous step are used in two separate fully-connected networks to estimate the probability of LC manoeuvres and the TTLC.
\end{enumerate}

Figure~\ref{whole_model} illustrates the proposed method and its key processing steps.

\subsection{BEV Input Data Representation} \label{in_rep}

Three binary layers of information are considered in forming the BEV representation: (1) The vehicle layer is populated by the 2D-bounding boxes, representing the vehicles at a single snapshot of the environment, (2) lane layer depicts the lane markings, and (3) road layer specifies the drivable area. The single-channel BEV representation is rendered by taking a layer-wise average of these three layers.Figure~\ref{qual_rlc_fig} and~\ref{qual_llc_fig} illustrates example of BEV representation in RLC and LLC scenarios. Two characteristics are considered in the BEV representation that facilitates the feature learning process. Firstly, unlike~\cite{Lee2017, Manttari2018} where the image is centred on the EV, we centre the BEV representation on the TV at each time step. This causes the position of lane markings in the representation to indicate the lateral position of the TV in the lane, which is one of the clear predictors of an LC manoeuvre. Centring the representation on the TV also allows encoding the relative states of the SVs compared to the TV which is informative in interaction modelling. Secondly, we use a lateral dimension resolution four times higher than the longitudinal dimension of the representation. This implies a magnification of lateral motions of vehicles in the representation which contains more informative features for the LC prediction problem. The size of the BEV representation in this study is considered to be 200 by 80 pixels covering 200 meters of the road in the longitudinal direction and 20 meters in the lateral direction. The 20 meters lateral coverage assures that the TV's adjacent lanes are always included. The input to the convolutional neural network is a multi-channel image of size $(T_{obs}*FPS)\times200\times80$, where each channel is a BEV representation rendered for a frame within observation window $T_{obs}$.


\begin{figure*}[!t]
\centering
\includegraphics[width=6.8in]{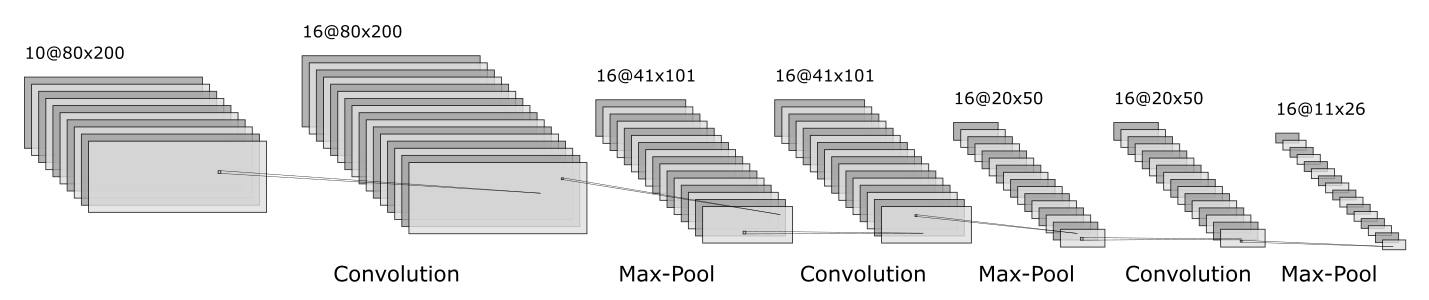}
\caption{The architecture of the CNN model.  }
\label{cnn}
\end{figure*}

\begin{figure}[!t]
\centering
\includegraphics[width=3.6in]{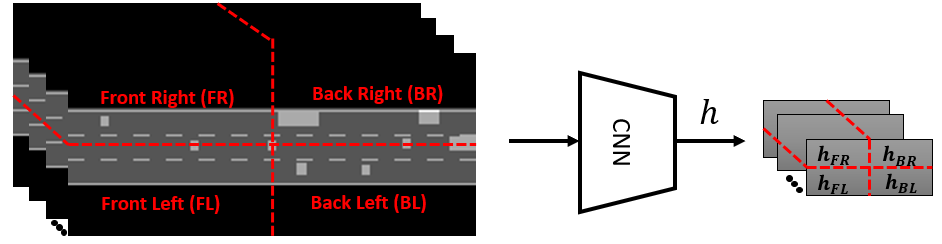}
\caption{Four areas of input representation and their corresponding areas in the CNN feature map. }
\label{att_fig}
\end{figure}

\subsection{Attention-based CNN for Feature Learning} \label{att_feature}
We propose an attention-based CNN to extract relevant spatio-temporal features from the temporally stacked BEV representation. Due to the sparsity of the BEV input representation, it is not required to use a very deep CNN, such as ResNet models~\cite{He2016}. Based on our empirical study, we consider a six-layer CNN including three convolution and three pooling layers. In each convolution layer, 16 learnable convolutional kernels are used with a size of $3*3$. The stride and padding of the kernels are set to 1 so that the size of the representation doesn't change after passing a convolutional layer. Each convolutional layer is followed by a Maximum Pooling (Max-Pool) layer with a size of $2*2$ to reduce the dimensionality of input data and a Rectified Linear Unit (ReLU) activation function. Figure~\ref{cnn} illustrates the architecture of the CNN model.

The spatial attention mechanism is proposed to enhance the performance of feature learning. The attention mechanism, inspired by the human brain, tries to selectively focus on a few relevant parts of the input data to estimate each output. In this study, the goal of employing an attention mechanism is to identify and focus on parts of the environment around the TV that have the most impact on the future behaviour of the TV. In a driving environment, normally only some part of the surrounding vehicles are contributing to the next manoeuvre of a vehicle. For example, a slow-moving vehicle in front usually leads to an RLC if there is a suitable gap in the right lane. In this example, the behaviour of SVs driving on the left lane does not influence the RLC decision made by the TV. Focusing on the relevant areas of the environment around the TV  is expected to increase the performance of LC prediction. Therefore, we divide the input representation into four areas, namely, 1) TV's Front Right (FR), 2) TV's Front Left (FL), 3) TV's Back Right(BR), and 4) TV's Back Left (BL). Since the input images are centred on the TV and vehicles are driving from right to left in input images, these four areas correspond to top-left, bottom-left, top-right, and bottom-right of the input images, respectively, as illustrated in Figure~\ref{att_fig}. The local connections in convolution and pooling layers of the CNN preserve the spatial location of the extracted features. Hence, the corresponding four areas in the CNN feature map, $h$ are selected as the hidden representations of the respective four areas in the TV's surrounding environment, namely, $h_{FR}$,$h_{FL}$ ,$h_{BR}$ , and $h_{BL}$ (see Figure~\ref{att_fig}). Note that these feature maps represents the TV's surrounding areas for all frames in the observation window. The attention weight for each area is estimated using a linear layer with softmax activation function as follows:
\begin{equation}
\alpha_{i} = Softmax(W\hat{h_i}+b),
\end{equation}
where $\hat{h_i}$ is the flattened vector of $h_i$. The attention weight for area $i$ is used to create a mask $m_i$ with the same size as $h$. $m_i$ is filled with values $\alpha_{i}$ in the regions corresponding to $h_i$ and zero elsewhere. The output of feature learning model is computed as the context vector $c$:
\begin{equation}
c = \sum_{i=1}^{n}m_{i} h,
\end{equation}

\subsection{Multi-task Prediction (MTL)}\label{formulation}

We leverage MTL in training our proposed prediction model for both future LC manoeuvre classification and TTLC regression tasks. Although it is possible to derive the classification information from TTLC estimates~\cite{Wirthmuller2021}, we adopt an MTL approach to benefit the generalisation gained by training the model on a similar parallel task. In addition, since the TTLC regression is more difficult than the three-class classification of future LC manoeuvres, the regressor performance can be enhanced by \textit{eavesdroping} on features learnt by the classifier.

Figure~\ref{whole_model} provides an overview of the proposed model. The output of the attention-based CNN feature extractor is shared among the classifier and the regressor sub-networks. Each of these sub-networks is a two-layer fully-connected neural network with ReLU activation function and dropout with a ratio of  $0.5$.
The classifier network has 128 hidden neurons with 3 outputs each corresponding to one of the LC manoeuvres.  These outputs are followed by a softmax activation function to decode their values as probabilities of LC manoeuvres. The regressor network has 512 hidden neurons with single output that specifies the predicted TTLC. A ReLU activation function is used at the output layer of the regression network as negative values are not accepted for the TTLC.  We consider a Cross-Entropy (CE) and Mean-Squared Error (MSE) loss functions for the classifier and regressor sub-networks as bellow:

\begin{equation}\label{ce_equ}
 L_{CE}= -\frac{1}{n} \sum_{i=1}^{n}\sum_{c=1}^{3} y_{i,c}\log{\hat{y_{i,c}}}
\end{equation}

\begin{equation}\label{mse_equ}
 L_{MSE} = \frac{1}{n} \sum_{i=1}^{n}(x^i-\hat{x^i})^2
\end{equation}
In the above equations, $n$ is the number of the training samples. $y_{i,c}$ and $\hat{y_{i,c}}$ are the ground truth and predicted probability of sample $i$ belonging to class $c$, respectively. The ground-truth and predicted TTLC for the sample $i$ are denoted as $x^i$ and $hat{x^i}$, respectively. The proposed multi-task model is trained based on the summation of the classifier and regressor losses as bellow:

\begin{equation}\label{loss_equ}
 L = L_{CE} + \gamma L_{MSE}
\end{equation}
where $\gamma$ defines the ratio between the regressor and classifier losses.

\begin{table}[]
\centering
\caption{Values of CL parameters during initial training epochs and remaining training epochs. Max included TTLC determines the maximum TTLC of included data samples in a epoch.}
\label{cl_table}
\begin{tabular}{@{}llllllll@{}}
\toprule
\multirow{2}{*}{}                                                        & \multicolumn{6}{l}{\textbf{Initial   Training Epochs}}                      & \textbf{\begin{tabular}[c]{@{}l@{}}Remaining\\  Training \\ Epochs\end{tabular}} \\
                                                                         & \textbf{0} & \textbf{1} & \textbf{2} & \textbf{3} & \textbf{4} & \textbf{5} & \textbf{\textgreater{}5}                                                         \\ \midrule
\textbf{\begin{tabular}[c]{@{}l@{}}Max \\ included \\ TTLC\end{tabular}} & 0.2        & 1.2        & 2.2        & 3.2        & 4.2        & 5.2        & 5.2                                                                              \\
\textbf{Loss Ratio}                                                      & 0          & 0.2        & 0.4        & 0.6        & 0.8        & 1          & 1                                                                                \\ \bottomrule
\end{tabular}
\end{table}

\subsection{Curriculum Learning (CL)}\label{cl}

We utilise curriculum learning based on two criteria specific to the LC prediction problem. First, predicting the LC manoeuvre in samples with smaller TTLC is generally easier compared to samples with larger TTLC. The reason is that as we get closer to the time when the TV crosses the lane marking, more explicit predictors can be found in the TV's motion. As the TTLC increase, the chance of mistakenly predicting an LC as an LK manoeuvre increases, until $TTLC>T_{pw}$ where the future manoeuvre is actually considered as LK. Therefore, we start training the network with samples with near-zero TTLC and gradually expose samples with larger TTLC to the prediction model. Second, predicting a class within a three-class classification task is normally considered easier than regressing a continuous variable. Hence, we start the training process by giving more importance to the classification task and gradually shift the focus to the regression task. This is achieved by increasing the loss ratio $\gamma$ in Equation~\ref{loss_equ} from 0 to 1 during the training phase. Table~\ref{cl_table} shows the TTLC and the loss ratio profile used during the training process.

\section{Performance Evaluation} \label{exps}
This section describes the evaluation of the proposed LC prediction method. First, the dataset and the LC scenario extraction steps are presented. Then, the implementation details of the proposed method are explained, followed by a discussion of the evaluation metrics. Next, the quantitative results comparing the proposed method with SOTA baseline models and the qualitative results are presented. Finally, an ablation study is performed on key components of our proposed method.

\begin{figure}[!t]
\centering
\includegraphics[width=3.4in]{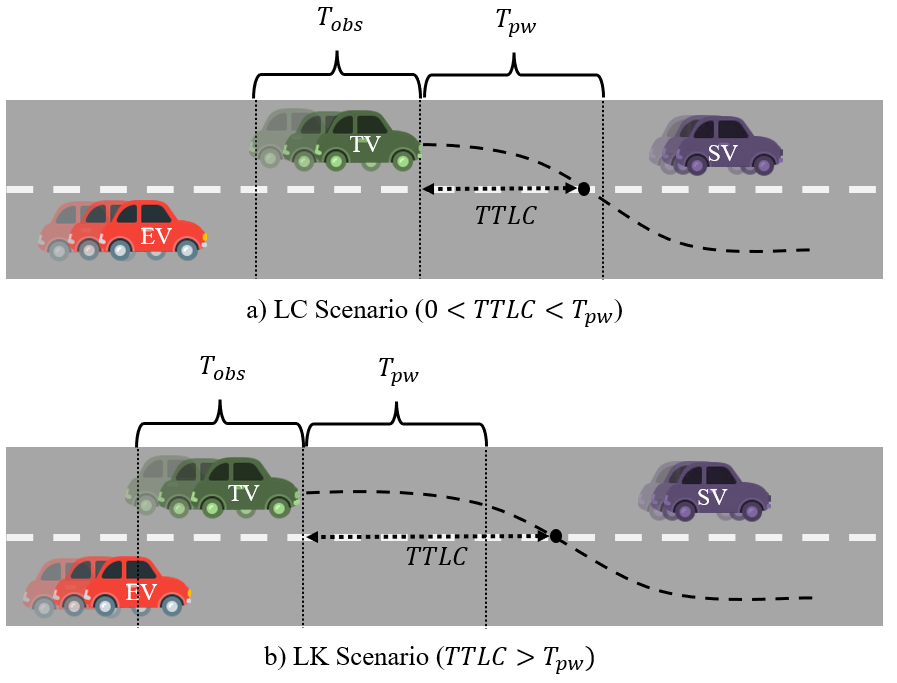}
\caption{ Examples of (a) RLC and (b) LK scenarios. In LK scenarios, the TV may change its lane after the $T_{pw}$ cut-off. }
\label{prob_def_fig}
\end{figure}

\subsection{Dataset and LC Scenario Extraction}\label{dataset}
The Highway Drone Dataset (highD)~\cite{highDdataset} consists of naturalistic vehicle trajectory data recorded at German highways using a drone. It contains the trajectories of 110500 vehicles observed over a highway segment of 420 meters at six different locations. We extract the LC and LK scenarios from the highD dataset to train and evaluate our proposed LC prediction method. An LC scenario is a group of data samples from a single vehicle with TTLC values ranging from $0$ to $T_{pw}$. Each data sample in an LC scenario at a given time $t_0$ contains the observation of the vehicles' trajectories in the driving environment during $[t_0-T_{obs},t_0-1]$. To have a balanced number of data samples per each value of TTLC, we do not consider LC scenarios where the corresponding trajectory data is not available for the full duration of $T_{obs}+T_{pw}~\mbox{sec}$ before the lane crossing. All the samples within an LC scenario are labelled as RLC or LLC according to the direction of the manoeuvre at the end of the scenario. An LK scenario is defined as a group of data samples from a single vehicle with $TTLC>T_{pw}$. Figure~\ref{prob_def_fig} illustrates an example of LC and LK scenarios. Similar to LC  scenarios, we consider $T_{pw}*\mbox{FPS}$ data samples within an LK scenario. In the highD dataset, the number of LK scenarios are much higher than LC scenarios. However, to keep the dataset balanced, we undersample the LK class so that the number of LK samples is equal to the average between the number of RLC and LLC samples. The challenging LK scenarios for training the prediction model are the ones that the TV changes its lane shortly after $T_{pw}$, since they can be easily misclassified as LC. Therefore, it is desirable to have more number of such challenging samples. However, there is not a sufficient number of such LK samples in the highD dataset due to the limited length of the covered road section. The highD dataset is published as 60 spreadsheets, from which we select the data in the first 50 files as training data, the next 5 files as validation data, and the remaining 5 files as test data. In total, we have extracted 7487, 932, and 698 LC/LK scenarios from training, validation, and test data, respectively.

\subsection{Implementation Details}\label{imp_details}
We train the proposed model and the re-implemented models from the literature on extracted LC/LK scenarios in the training set of the highD dataset. To train our proposed prediction model, Adam optimiser~\cite{kingma2017adam} is used with a learning rate of $0.001$ and a maximum of 20 training epochs. The early-stopping technique on validation loss function is used to avoid over-fitting. All the models have been implemented using the PyTorch framework~\cite{paszke2019pytorch} and are trained and evaluated on a single GeForce RTX 2080 Ti GPU. We use a batch size of 64 during the training period. The source code of this study is available at \url{https://github.com/SajjadMzf/EarlyLCPred}.

\subsection{Evaluation Metrics}\label{metrics}
In this subsection we discuss the metrics we used in evaluation of the future LC classifier and TTLC regressor.

\begin{table}[]
\centering
\begin{threeparttable}
\centering
\caption{Confusion Matrix Used For LC Prediction}
\label{confusion_mat}

\begin{tabular}{@{}cccc@{}}
\toprule
                \textbf{Ground-Truth\textbackslash Predicted}    & \textbf{ LK} & \textbf{ RLC} & \textbf{ LLC} \\ \midrule
\textbf{ LK}  & TN*                    & FP                     & FP                     \\
\textbf{ RLC} & FN                    & TP                     & FN, FP                 \\
\textbf{ LLC} & FN                    & FN, FP                 & TP                     \\ \bottomrule
\end{tabular}
\begin{tablenotes}
      \small
      \item * TN: True Negative, TP: True Positive, FP: False Positive, FN: False Negative 
    \end{tablenotes}
\end{threeparttable}
\end{table}

\subsubsection{Classification Metrics}

Similar to previous studies, we report accuracy, precision, recall and F1 score over all data samples in the test dataset. However, relying on such metrics is not sufficient for a comprehensive evaluation of a prediction model, especially for applications in automated driving. A reliable LC predictor is expected to have high recall particularly in samples with short TTLC since missing such LC samples can create safety-related issues for the EV and the surrounding traffic. To this end, we measure the recall (a.k.a. True Positive Rate) of the prediction model with respect to the actual TTLC of the data samples. 

Precision, recall, and F1 score are usually defined for a binary classifier. To extend them to a multi-class problem, one can use the "one-vs-all" approach. However, we consider both the RLC and LLC classes as positive classes (similar to~\cite{Rehder2019}), since both are the rarely occurring manoeuvres that we are interested in predicting. Therefore, we consider the confusion matrix for the LC prediction problem as Table~\ref{confusion_mat}. Note that the data samples with RLC or LLC labels which are classified incorrectly in LLC or RLC classes, respectively, are considered in both FN and FP. The reason is that in such cases the prediction model misses the prediction of the type of LC and generates an incorrect LC alarm.

In addition to having a high recall, an LC prediction model is expected to have a low rate of false LC alarms (False Positive). Although False alarms may not create safety issues, they decrease the comfortability of the driving system because the high rate of false alarms causes frequent, unnecessary slow-downs by the EV. The trade-off between the False Positive Rate (FPR) and True Positive Rate (TPR) of a classifier is reported in terms of the Receiver Operating Characteristic (ROC) Curve and the Area Under the ROC Curve (AUC) of the prediction model. 


To measure the average prediction horizon of a prediction model, we report the following two metrics, similar to~\cite{Wirthmuller2020J}.
\begin{itemize}
\item First prediction time, denoted by $\tau_f$ : is the time between the first correct prediction of the manoeuvre class and the $T_{cross}$, averaged on all data samples.
\item Robust prediction time, denoted by $\tau_c$: is the time between the first moment when the model starts to continuously predict correctly and the $T_{cross}$, averaged on all data samples..
\end{itemize}

\subsubsection{Regression Metrics} 

Although we use the MSE loss function in the training phase, we adopt the RMSE metric to evaluate the performance of the TTLC prediction. The reason is that RMSE is more understandable than MSE since it has the same physical unit as the output data (i.e., seconds). To evaluate the performance of the prediction model for different TTLCs, we use the box plot of the predicted TTLC for each actual TTLC, specifying the median values of the predicted TTLC.

\begin{table*}[]
\centering
\caption{Comparison of the proposed LC prediction model with SOTA on the test set of HighD dataset}
\label{comp_table}
\begin{tabular}{@{}llcccccccc@{}}
\toprule
Task                            & Model                        & Accuracy      & Recall        & Precision     & F1-score      & AUC           & $\tau_f$      & $\tau_c$      & RMSE           \\ \midrule
\multirow{5}{*}{Classification} & MLP1~\cite{Wirthmuller2020J} & 0.75          & 0.65          & \textbf{0.94} & 0.77          & 0.84          & 3.97          & 2.73          & -              \\
                                & MLP2~\cite{Shou2020}         & 0.59          & 0.52          & 0.74          & 0.61          & 0.61          & 3.49          & 2.02          & -              \\
                                & LSTM1~\cite{Wirthmuller2021} & 0.79          & \textbf{0.90} & 0.75          & 0.82          & 0.86          & 4.24          & 2.98          & -              \\
                                & LSTM2                        & 0.78          & 0.84          & 0.81          & 0.82          & 0.84          & 4.43          & 3.76          & -              \\
                                & CS-LSTM~\cite{Deo2018}       & 0.74          & 0.78          & 0.81          & 0.72          & 0.76          & 3.92          & 3.61          & -              \\
\multirow{2}{*}{Regression} \rule{0pt}{4ex}     & LSTM1~\cite{Wirthmuller2021} & -             & -             & -             & -             & -             & -             & -             & 0.841          \\
                                & LSTM2                        & -             & -             & -             & -             & -             & -             & -             & 0.976          \\
Dual                            & Proposed\rule{0pt}{4ex}                     & \textbf{0.83} & 0.85          & 0.85          & \textbf{0.85} & \textbf{0.88} & \textbf{4.75} & \textbf{3.96} & \textbf{0.629} \\ \bottomrule
\end{tabular}
\end{table*}

\begin{table*}[]
\centering
\caption{List of features for baseline models in comparitive study}
\label{feature_table}
\begin{tabular}{@{}ll@{}}
\toprule
Baseline                     & List of Features                                                                                                                                                                                                                                                                                                                                                                                                                                                                                                                                                                                                                                                                                                                                                                                                                                     \\ \midrule
MLP1~\cite{Wirthmuller2020J} & \begin{tabular}[c]{@{}l@{}}(1) Existence of left lane, (2) Existence of right lane, (3) Lane width, (4) Longitudinal distance of TV to PV,\\ (5) Longitudinal distance of TV to RPV, (6) Longitudinal distance of TV to FV, \\ (7) Lateral distance of TV to the left lane marking, (8) Lateral distance of TV to RV, (9) Lateral distance of TV to RFV,\\ (10) Relative longitudinal velocity of TV w.r.t. PV, (11) relative longitudinal velocity of TV w.r.t. FV, \\ (12) Relative lateral velocity of TV w.r.t. PV, (13) Relative lateral velocity of TV w.r.t. RPV,\\ (14) Relative lateral velocity of TV w.r.t. RV, (15) Relative lateral velocity of TV w.r.t. LV,\\ (16) Longitudinal acceleration of the TV, (17)Relative longitudinal acceleration of the TV w.r.t RPV,\\ (18) Lateral acceleration of the prediction target\end{tabular} \\
MLP2~\cite{Shou2020}         & \begin{tabular}[c]{@{}l@{}}(1) Existence of left lane, (2) Existence of right lane, (3) Longitudinal distance of TV to RPV, \\ (4) Longitudinal distance of TV to PV, (5) Longitudinal distance of TV to LPV, (6) Longitudinal distance of TV to RV,\\ (7) Longitudinal distance of TV to LV, (8) Longitudinal distance of TV to RFV, (9) Longitudinal distance of TV to FV, \\ (10) Longitudinal distance of TV to LFV, (11) Relative velocity of TV w.r.t. RPV, (12) Relative velocity of TV w.r.t. PV,\\ (13) Relative velocity of TV w.r.t. LPV, (14) Relative velocity of TV w.r.t. RV, (15) Relative velocity of TV w.r.t. LV, \\ (16) Relative velocity of TV w.r.t. RFV, (17) Relative velocity of TV w.r.t. FV, (18) Relative velocity of TV w.r.t. LFV\end{tabular}                                                                        \\
LSTM1~\cite{Wirthmuller2021} & Same as MLP1~\cite{Wirthmuller2020J}                                                                                                                                                                                                                                                                                                                                                                                                                                                                                                                                                                                                                                                                                                                                                                                                                 \\
LSTM2                        & \begin{tabular}[c]{@{}l@{}}(1) Lateral velocity, (2)   Longitudinal velocity, (3) Lateral acceleration, (4) Longitudinal   acceleration, \\ (5) Lateral distance of TV to the left lane marking, (6)   Relative longitudinal velocity of the TV w.r.t. PV,\\  (7) Longitudinal distance   of TV to PV, (8) Relative longitudinal velocity of the TV w.r.t. FV, \\ (9)   Longitudinal distance of TV to FV, (10) Longitudinal distance of TV to RPV,   (11) Longitudinal distance of TV to RV,\\  (12) Longitudinal distance of TV to   RFV, (13) Longitudinal distance of TV to LPV, \\ (14) Longitudinal distance of   TV to LV, (15) Longitudinal distance of TV to LFV, (16) Existence of left   lane, \\ (17) Existence of right lane, (18) Lane width\end{tabular}                                                                              \\ \bottomrule
\end{tabular}
\end{table*}

\subsection{Quantitative Results} \label{quan_results}
The results of an LC prediction model depends on the selected dataset and preprocessing steps used for training and evaluation of the model. Factors such as the size of the dataset, the level of noisiness, traffic density, and LC scenario extraction steps can impact the prediction model's results. In addition, parameters of the problem definition can influence the overall performance reported for a prediction model. For example, selecting a smaller prediction window $T_{pw}$ results in better overall prediction performance since it is generally easier to predict manoeuvres with shorter TTLC. To do a fair comparison among our proposed model and the existing solutions, we re-train and evaluate some baseline models from the literature on the same dataset, using the same preprocessing steps and parameters of the problem definition used in this study. We select the following SOTA approaches, where an implementation code or details of implementation are provided:

\begin{figure}[!t]
\centering
\includegraphics[width=3.4in]{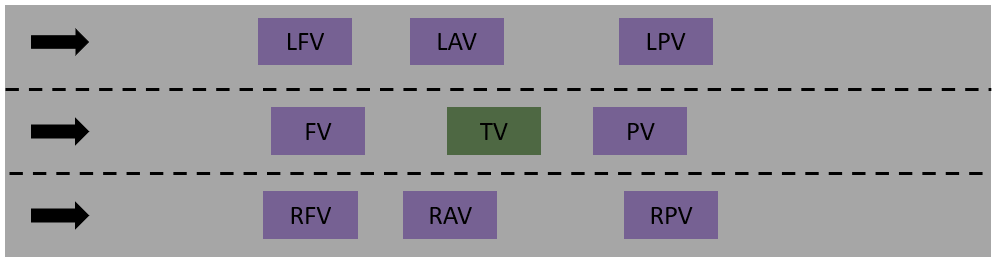}
\caption{Terminology of the TV's surrounding vehicle}
\label{sv_names}
\end{figure}

\begin{figure}[!t]
\centering
\includegraphics[width=3.4in]{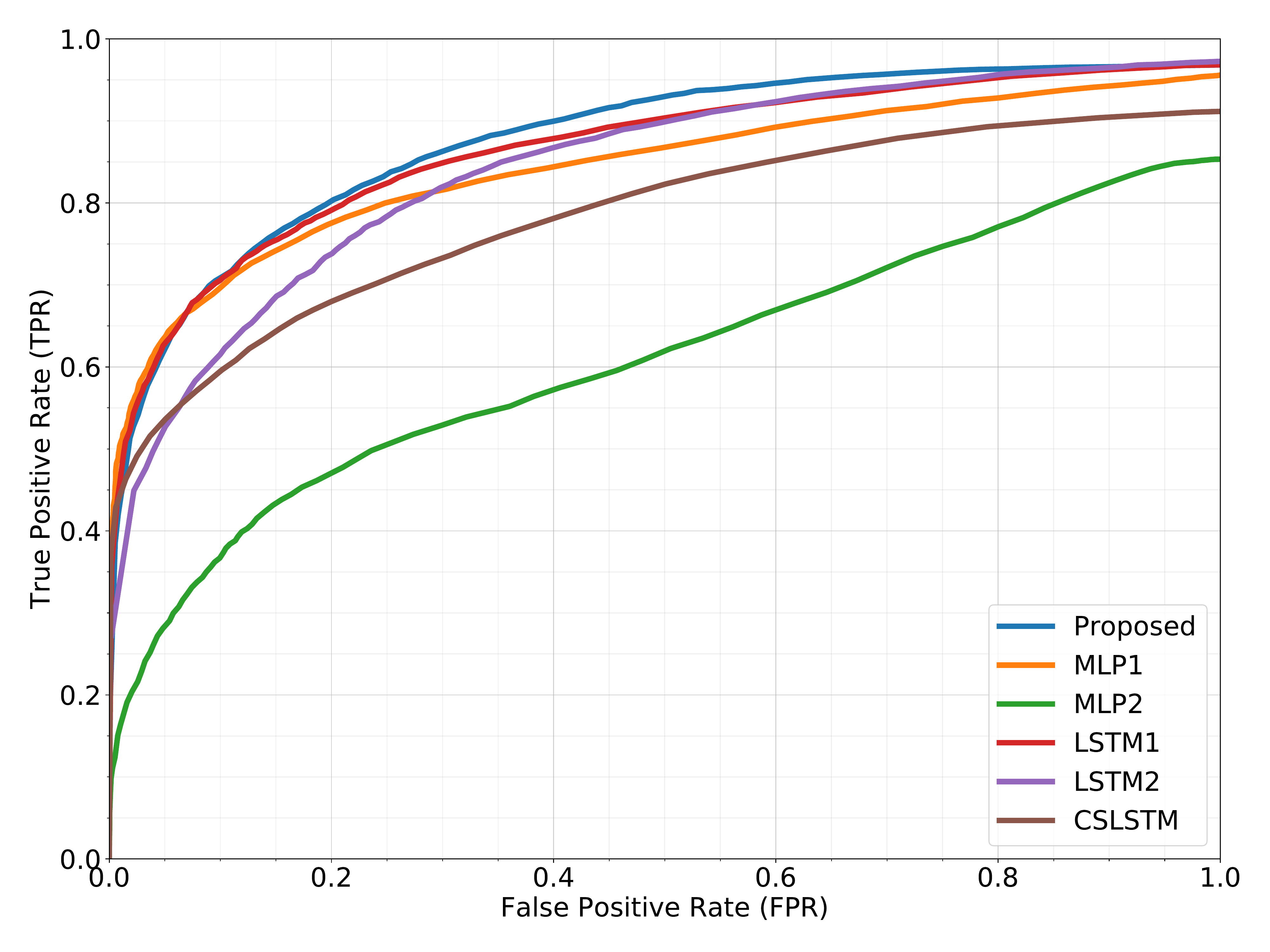}
\caption{Receiver Operating Characteristics (ROC) Curve of the proposed method and SOTA on highD dataset with prediction window $T_{pw}$ of 5 seconds.  }
\label{roc_fig}
\end{figure}

\begin{itemize}
\item \textbf{MLP}: Multi-Layer Perceptrons (MLPs) have been used in some existing studies~\cite{Yoon2016, Kruger2019, Wirthmuller2020J, Shou2020}. Here, we re-implement a two-layer MLP with 512 hidden neurons and a ReLU activation function. We consider two sets of input features for the MLP model:
\begin{enumerate}
\item \textbf{MLP1}: In~\cite{Wirthmuller2020C, Wirthmuller2021}, a study has been performed on a large-scale dataset to select informative features for MLP and LSTM networks (see Table V in~\cite{Wirthmuller2021}). The study in~\cite{Wirthmuller2021} provides 21 features from which we select 18. The authors did not provide the details of how to extract the other 3 features from highD dataset.
\item \textbf{MLP2}: In~\cite{Shou2020}, MLPs are used to predict left lane changes. The authors argue that not including lateral distance features results in better performance in very long prediction horizons ($>5$sec). In~\cite{Shou2020}, they have only considered relative features of the TV w.r.t.  preceding vehicle, left preceding vehicle (PV), and left following vehicle (LFV). In our comparison, the same set of relative features is used also for the right preceding vehicle (RPV) and right following vehicle (RFV) to extend their study to both right and left lane change prediction. 
\end{enumerate}

\item \textbf{LSTM}: Several existing studies on LC prediction~\cite{Scheel2019, Ding2019, Wirthmuller2021} have used LSTM due to its power in processing data sequences. A single-layer LSTM model is used with a hidden state size of 512 followed by a two-layer fully connected neural network with a hidden state of 128 and 512 for classification and regression tasks, respectively. Two different set of input features are used for the LSTM model:

\begin{enumerate}
\item  \textbf{LSTM1}: Similar to the MLP model, we adopt the feature list used in~\cite{Wirthmuller2021}.
\item \textbf{LSTM2}: The rationale behind this feature set is to: (1)describe TV's motion (Features 1 to 5), (2) detect slow-moving preceding vehicle (Features 6 and 7), (3) high-speed following vehicle (Features 8 and 9), and (4) detect the gap in right/left lane (Features 10 to 18)
\end{enumerate}

\item \textbf{CS-LSTM}: Convolutional social pooling LSTM developed by Deo and Trivedi~\cite{Deo2018} combines the power of CNNs in spatial interaction modelling with the power of LSTMs in modelling dynamics of each vehicle. The code published by the authors is used to re-train and evaluate their model with our preprocessed dataset and the parameters used in our problem definition.
\end{itemize}

\begin{figure}[!t]
\centering
\includegraphics[width=3.4in]{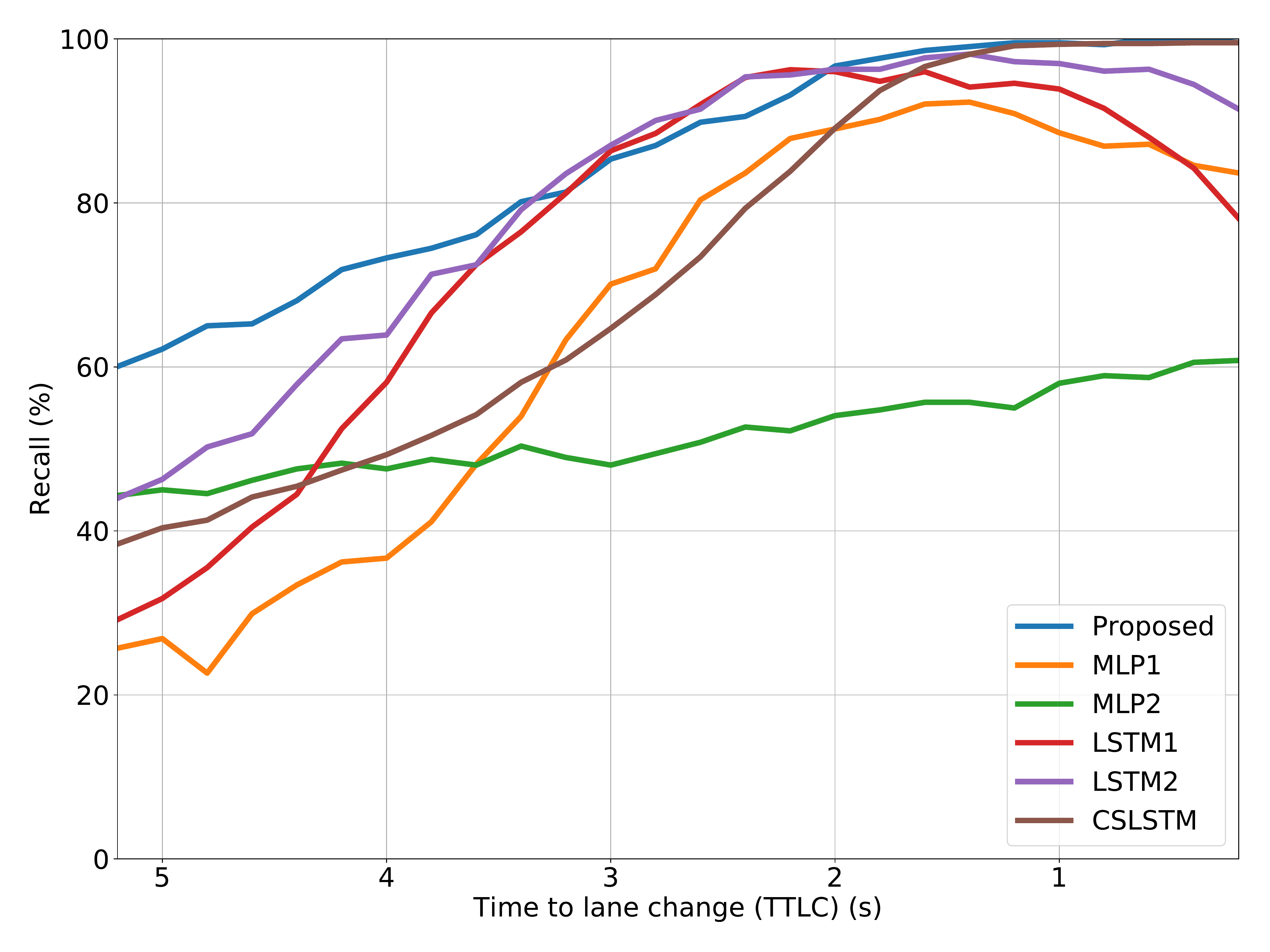}
\caption{Percentages of Recall vs TTLC of the proposed method and SOTA on the test set of highD dataset }
\label{recall_fig}
\end{figure}

Table~\ref{feature_table} contains the lists of features used for each baseline prediction model. The terminology on the TV's surrounding vehicles used in the feature lists is illustrated in Figure~\ref{sv_names}. Table~\ref{comp_table} shows the results of the selected models for both classification and regression tasks and the results of the proposed model in the dual-task. In the classification task, the proposed model outperforms SOTA approaches with an increase of $4\%$ in accuracy and $3\%$ in F1-score. The LSTM1 and LSTM2 models achieve the second-best performance in classification with $0.82$ F1-score. Although the MLP1 and LSTM1 models achieve best precision and recall, respectively, in total they achieve lower F1-scores compared to proposed approach.  In regard to the prediction horizon, the LSTM2 model ranks first among selected models from the literature by 3.76 seconds robust prediction time. Nevertheless, the proposed model outperforms CSLSTM by 0.3 seconds in the robust prediction time metric. In the regression task, the proposed model improves the RMSE metric by around 0.2 seconds compared to the LSTM network used in~\cite{Wirthmuller2021}.

To evaluate the LC prediction performance at different likelihood thresholds, we plotted the ROC curve in Figure~\ref{roc_fig}. The area under the curve (AUC) is reported in Table~\ref{comp_table}. The proposed method outperforms SOTA by around $2\%$ in AUC and has a higher True Positive Rate (TPR) for different thresholds compared to SOTA. Note that even with $100\%$ False Positive Rate (FPR), none of the models achieves $100\%$ TPR. The reason is that there still might be some samples within the positive classes that are classified in wrong positive classes (i.e., RLC data classified as LLC data and vice versa). As we mentioned earlier such samples are reported as both FP and FN of the prediction model. The proposed model, compared to the reported SOTA models, reduces the number of such samples by more than half, as indicated in Figure~\ref{roc_fig}.

Figure~\ref{recall_fig} reports the recall (a.k.a. TPR) vs the TTLC. Both the proposed and the CSLSTM models have close to $100\%$ recall in predictions with TTLC less than 1.5 seconds, while the MLPs and LSTMs miss at least $10\%$ of LC manoeuvres occurring in less than 1.5 seconds. As discussed previously, any missed short-term predictions can lead to safety-related issues. In terms of long term prediction, the proposed method significantly outperforms the SOTA methods. For instance, in samples with TTLC equals 5.2 seconds, which is the maximum prediction horizon, the proposed method achieves $60\%$ recall which is an improvement of around $50\%$ compared to the SOTA models. Note that improving the performance of long-term predictions is challenging because generally there is no explicit change in the TV's behaviour for LCs that occur far in the future (i.e., more than 3 seconds). Therefore, the prediction model needs to infer the occurrence of an LC by extracting clues from the traffic context around the TV, which makes long-term prediction more challenging. The MLP1~\cite{Shou2020}, which relies only on longitudinal features,achieves comparative long-term prediction performance compared to other baseline models, while performs poorly in short-term prediction. This can be because of not using lateral features which contains clear clues for short-term lane change predictions.

\begin{figure}[!t]
\centering
\includegraphics[width=3.4in]{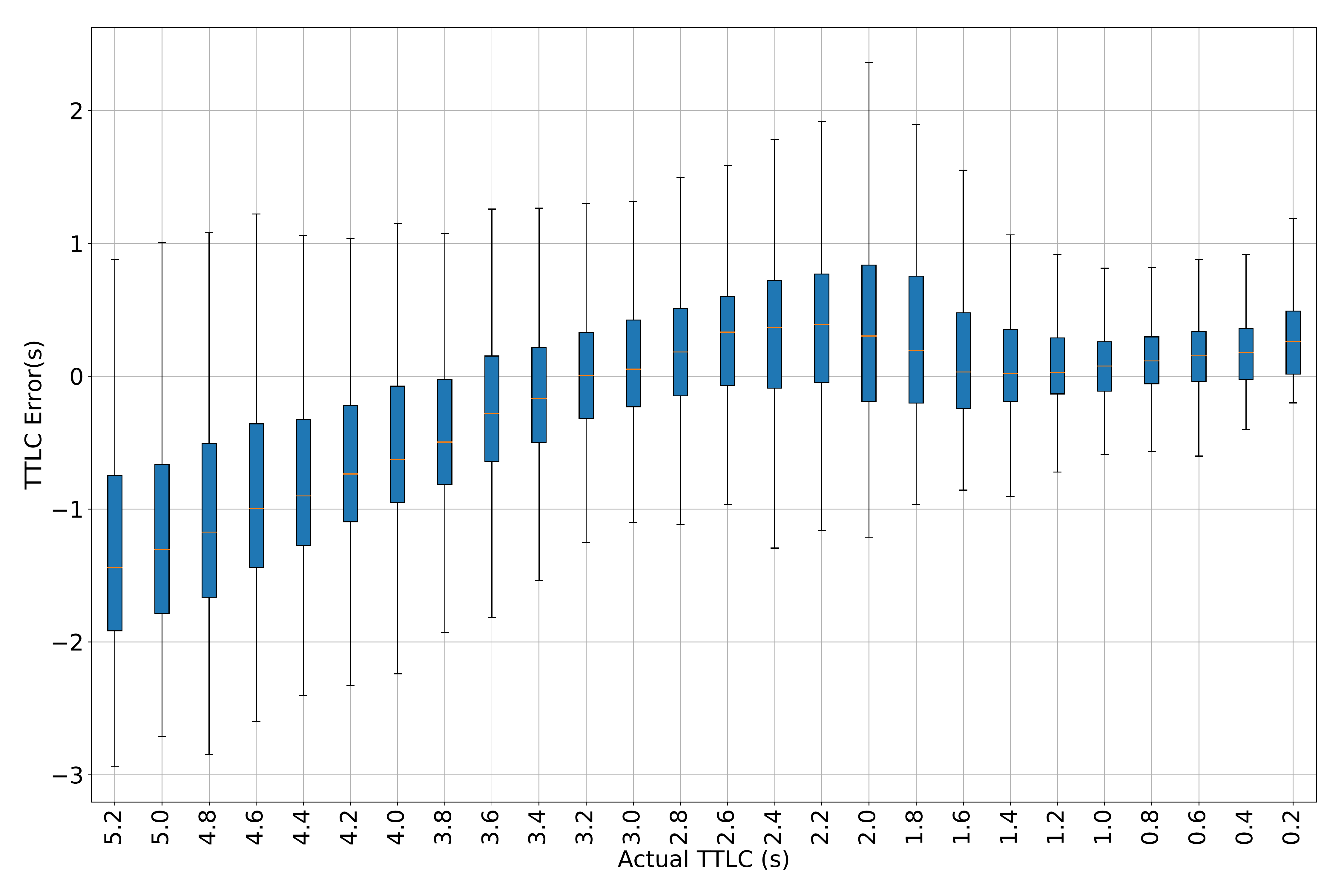}
\caption{ Box plot of the TTLC predictions error ($TTLC_{predicted}-TTLC_{ground truth}$) on test data of highD}
\label{box_fig}
\end{figure}

To evaluate the TTLC regression performance, we report the box plot for all TTLC prediction errors on the test dataset in Figure~\ref{box_fig}. The results show that samples with a higher TTLC generate predictions with higher median error and variance. In addition, for TTLC greater than 3.2 seconds the model tends to predict TTLCs less than actual values of TTLC with a median of error greater than 0.5 seconds. This can be explained by the fact that samples with TTLC greater than 3 seconds do not exhibit any explicit change in lateral movement of the TV. Therefore, it is not possible to estimate the lane crossing time using the information in lateral speed and distance to lane marking.

\subsection{Qualitative Results} \label{qual_results}

\begin{figure*}[!t]
\centering
\includegraphics[width=6.8in]{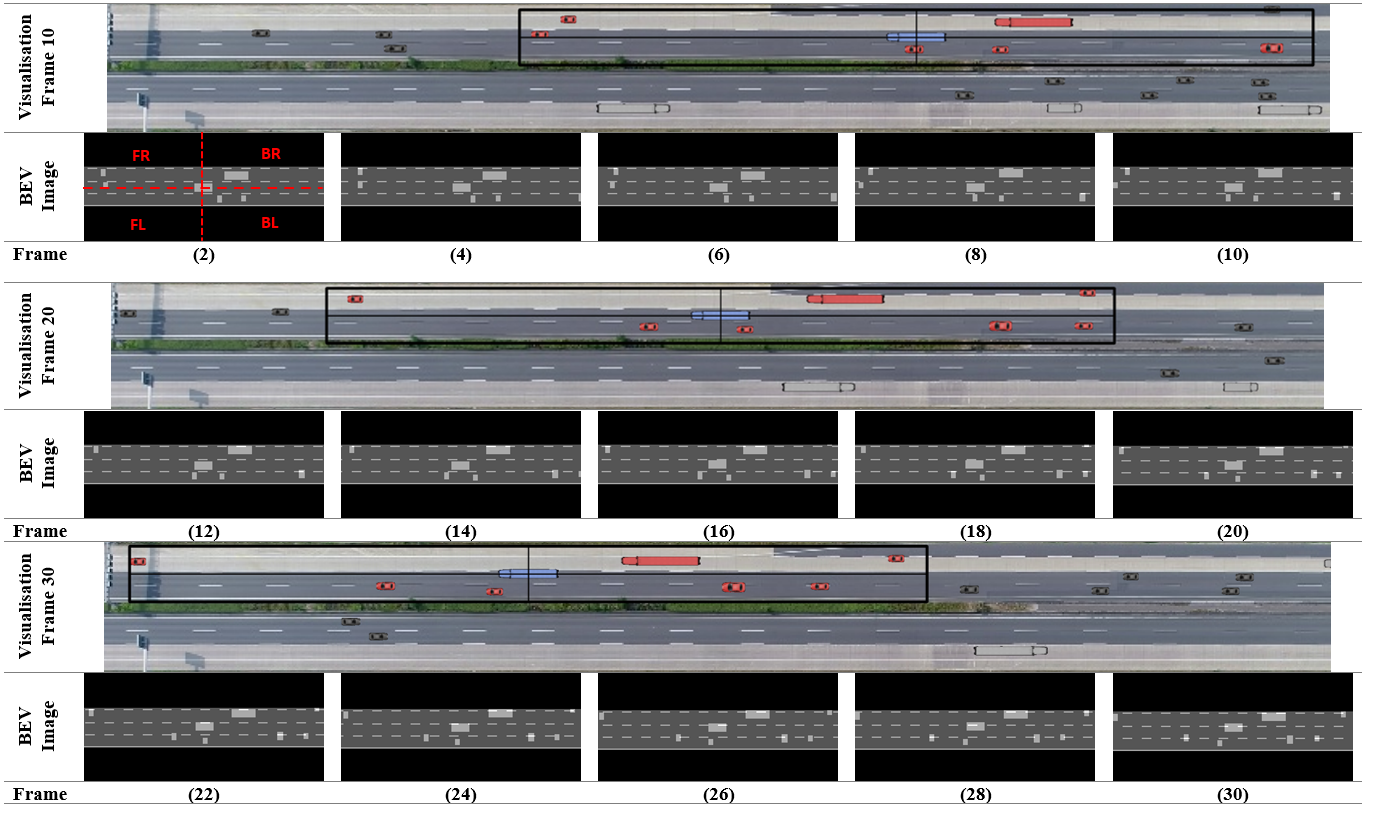}
\caption{ An example of RLC scenario. Four attention areas are specified in BEV image at frame 2 as an example. In visualisations at frame 10,20, and 30, the TV is depicted with blue, the vehicles within the cropped BEV image are depicted with red, and remaining vehicles are depicted with gray colours.}
\label{qual_rlc_fig}
\end{figure*}

\begin{figure*}[!t]
\centering
\includegraphics[width=6.8in]{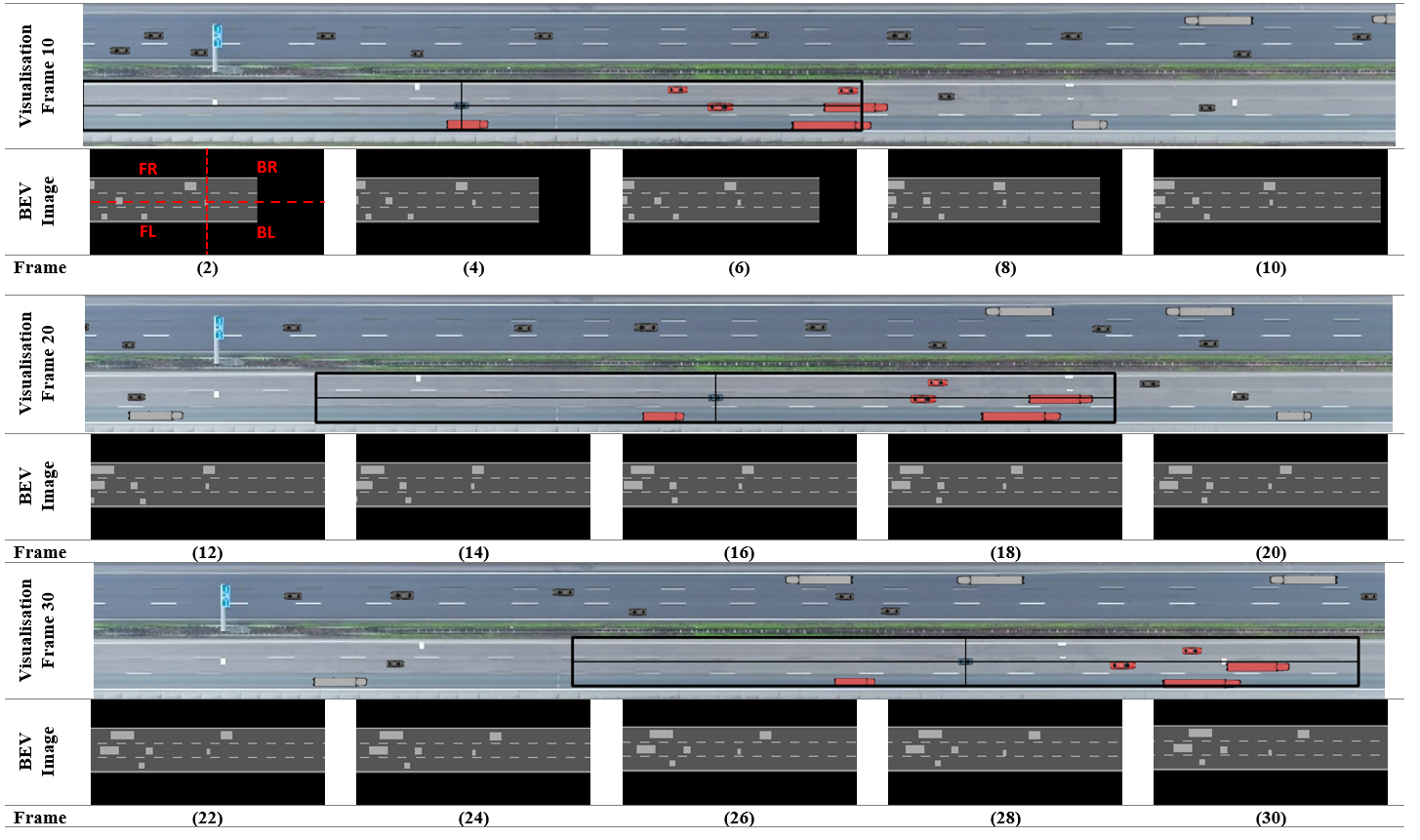}
\caption{ An example of LLC scenario.  Four attention areas are specified in BEV image at frame 2 as an example. In visualisations at frame 10,20, and 30, the TV is depicted with blue, the vehicles within the cropped BEV image are depicted with red, and remaining vehicles are depicted with gray colours.}
\label{qual_llc_fig}
\end{figure*}

Two examples of RLC and LLC scenarios are represented in Figure~\ref{qual_rlc_fig} and~\ref{qual_llc_fig}. The model's predictions and estimated attention weights in frame 10, 20, and 30 of these example scenarios are provided in Table~\ref{qual_rlc_table} and~\ref{qual_llc_table}. 

In Figure~\ref{qual_rlc_fig}, the TV (i.e., the blue truck) is going to complete an overtake of a slow-moving truck on the right lane (i.e., the red truck) by performing an RLC manoeuvre. At frame 10, the model predicts a $50\%$ chance for LK in the next 5.2 seconds and a $48\%$ chance for an RLC. The TV starts to move towards the right lane marking at frame 16. The prediction performance at frame 20 shows that the estimated likelihood of an RLC manoeuvre is increased to $85\%$. The clues of such a manoeuvre can be found in the available gap in the right lane in front of the slow-moving truck. At frame 10, where the prediction model gives almost equal possibilities to the LK and RLC manoeuvres, the most attention is given almost equally to the front right and back right of the TV. However, as the model becomes more confident in predicting the RLC manoeuvre the attention mechanism increases its focus on the back right area, where the slow-moving vehicle is located.

Figure~\ref{qual_llc_fig} shows the emergence of an LLC manoeuvre by the TV due to a slow-moving vehicle in front of the TV. In frame 10, the model confidently predicts the LLC manoeuvre 5.2 seconds before crossing the left lane marking, despite no lateral movement of the TV within frames 0 to 10. The model continues to predict the LLC manoeuvre correctly as the TV gets closer to the slow-moving proceeding vehicle and finally starts to move toward the left lane markings. The focus of the attention mechanism during these frames are on the traffic on the left lane, with more focus on the front left area. This could be explained by the higher information in the left lane for the predicted LLC manoeuvre. The video uploaded in~\cite{our_video} explains more examples of LK and LC scenarios and the model performance.

\begin{table*}[]
\centering
\caption{Prediction performance and attention weights for the example RLC scenario in figure~\ref{qual_rlc_fig}}
\label{qual_rlc_table}
\begin{tabular}{@{}llllllllll@{}}
\toprule
Frame & Grount Truth TTLC & Predicted TTLC & P(m=LK) & P(m=RLC) & P(m=LLC) & $\alpha_{FR}$ & $\alpha_{FL}$ & $\alpha_{BR}$ & $\alpha_{BL}$ \\ \midrule
10    & 5.2               & -              & 0.5     & 0.48     & 0.02     & 0.32          & 0.1           & 0.34          & 0.23          \\
20    & 3.2               & 3.05           & 0.15    & 0.85     & 0        & 0.2           & 0.14          & 0.46          & 0.2           \\
30    & 1.2               & 1.12           & 0       & 1        & 0        & 0.02          & 0.07          & 0.88          & 0.03          \\ \bottomrule
\end{tabular}
\end{table*}

\begin{table*}[]
\centering
\caption{Prediction performance and attention weights for the example LLC scenario in figure~\ref{qual_llc_fig}
}
\label{qual_llc_table}
\begin{tabular}{@{}llllllllll@{}}
\toprule
Frame & Grount Truth TTLC & Predicted TTLC & P(m=LK) & P(m=RLC) & P(m=LLC) & $\alpha_{FR}$ & $\alpha_{FL}$ & $\alpha_{BR}$ & $\alpha_{BL}$ \\ \midrule
10    & 5.2               & 4.87           & 0.03    & 0        & 0.97     & 0.07        & 0.73        & 0.06        & 0.14        \\
20    & 3.2               & 3.3            & 0       & 0        & 1        & 0.02        & 0.62        & 0.04        & 0.32        \\
30    & 1.2               & 0.98           & 0       & 0        & 1        & 0           & 0.94        & 0.05        & 0.01        \\ \bottomrule
\end{tabular}
\end{table*}

\subsection{Ablation Study} \label{ab_study}
In this subsection, we investigate the impact of the key components of the proposed prediction model by evaluating the model performance with different combination of these components. We use the AUC and RMSE on the validation data as evaluation metrics of classification and regression tasks, respectively. 

Table~\ref{ab_study_table} shows the results of the ablation study on Multi-Task Learning (MTL), Attention Mechanism, and Curriculum learning based on MTL loss and TTLC order. In this study, the CNN model (see Figure~\ref{cnn}) is used as a baseline and we gradually add the components to investigate their impact in both future LC classification and TTLC regression tasks. The results in Table~\ref{ab_study_table} shows that using the attention mechanism can enhance the AUC of the classifier by around $3\%$ and slightly decrease the error of the regressor. On the other hand, the MTL when used without attention and curriculum learning, can degrade the performance of the classification task and does not provide improvements for the regression task. However, the MTL is capable of improving the results and achieving the best performance when combined with other key components of the prediction model.

\section{Conclusion} \label{conclusion}
This paper proposed a multi-task attention-based CNN model for the early prediction of LC manoeuvres in highway driving scenarios. The prediction model was trained and evaluated using a large-scale naturalistic trajectory dataset. The results shows that the attention-based CNN is capable of extracting interaction-aware features from the surrounding traffic required for long-term prediction. The multi-task approach, boosted by two novel curriculum learning schemes, enables TTLC and manoeuvre likelihood prediction using shared extracted features. Furthermore, the proposed model outperforms selected SOTA LC prediction approaches, demonstrating a reliable short-term prediction and 1.5 times better long-term prediction performances.

In this study, we trained and tested our model on a naturalistic trajectory dataset recorded using a camera installed on a drone. Such a camera provides a wide and unobstructed view of the environment, while the on-board perception of automated vehicles is impaired with occlusion, sensor noise and limited field of view, which can create uncertainties in observing surrounding vehicles. Our future research will focus on quantifying the impact of these input uncertainties on the lane change prediction problem.

\begin{table}[]
\begin{threeparttable}
\centering
\caption{Results of Ablation studies on key components of the proposed method using the validation set of the highD dataset}
\label{ab_study_table}
\begin{tabular}{@{}cccccc@{}}
\toprule
Task        & Attention  & CL (Loss)  & CL (TTLC)  & AUC\%           & RMSE (s)       \\ \midrule
C*         &                  &            &            & 88.38          & -              \\
R**         &                  &            &            & -              & 0.804          \\
MTL &                  &            &            & 83.11          & 0.805          \\
MTL & \checkmark       &            &            & 86.89          & 0.796          \\
MTL & \checkmark       &            & \checkmark  & 87.79          & 0.809          \\
MTL & \checkmark       &  \checkmark &           & 87.97          & \textbf{0.774} \\
MTL & \checkmark       & \checkmark & \checkmark & \textbf{89.43} & \textbf{0.774} \\ \bottomrule
\end{tabular}
\begin{tablenotes}
      \small
      \item * C: Classification, ** R: Regression
    \end{tablenotes}
\end{threeparttable}
\end{table}

\ifCLASSOPTIONcaptionsoff
  \newpage
\fi

\bibliographystyle{IEEEtran}

\bibliography{references}

\end{document}